\documentclass[runningheads]{llncs}
\usepackage[T1]{fontenc}
\usepackage{graphicx}
\usepackage{color}
\usepackage{url}

\usepackage{hyperref}
\usepackage{booktabs}
\usepackage{array} 
\usepackage{tabularx}
\usepackage{arydshln}
\usepackage{subcaption}
\usepackage{amsmath}
\usepackage{cleveref}
\usepackage{amssymb}
\usepackage{xcolor}
\usepackage{subcaption}
\usepackage{tabularray}

\newcommand{\snum}[2]{$\textstyle #1 \pm \scriptstyle #2$}
\newcommand{\NA}{\text{\scalebox{0.85}{N.A.}}}

\newcommand{\snumbf}[2]{$\textstyle \mathbf{#1} \pm \scriptstyle \mathbf{#2}$}

\usepackage{xspace}

\usepackage{placeins}
\usepackage{stfloats} 
\usepackage[table]{xcolor}

\begin{document}
\title{How to Evaluate and Refine your CAM}
\author{Luca Domeniconi \and Alessandra Stramiglio \and Michele Lombardi \and Samuele Salti}
\authorrunning{L. Domeniconi et al.}
\institute{University of Bologna, Bologna, Italy\\
\email{luca.domeniconi5@studio.unibo.it\\ \{a.stramiglio, michele.lombardi2, samuele.salti\}@unibo.it}}
\maketitle

\begin{abstract}

Class attribution maps (CAMs) provide local explanations for the decisions of convolutional neural networks. While widely used in practice, the evaluation of CAMs remains challenging due to the lack of ground-truth explanations, making it difficult to evaluate the soundness of existing metrics. Independently, most commonly used CAM methods produce low-resolution attribution maps, which limits their usefulness for detailed interpretability.

To address the evaluation challenge, we introduce a synthetic dataset with ground-truth attributions that enables a rigorous comparison of CAM evaluation metrics. Using this dataset, we analyze existing metrics and propose ARCC, a new composite metric that more reliably identifies faithful explanations. To address the low resolution issue, we introduce RefineCAM, a method that produces high-resolution attribution maps by aggregating CAMs across multiple network layers. Our results show that RefineCAM consistently outperforms existing methods according to the proposed evaluation.

\end{abstract}

\section{Introduction}
\label{sec:intro}
In the context of computer vision, one of the most popular approaches to explain a single model prediction is to produce attribution maps \cite{molnar2020interpretable}. These maps aim to highlight the most influential pixels on the model's output. To produce them, two main approaches have been developed. The first is based on the gradients of activations in the network with respect to the input values \cite{GradCAM++,GradCAM,simonyan2013deep,sundararajan2017axiomatic}. The second is based on input perturbations and game theory \cite{petsiuk2018rise,fong2017interpretable}.\begingroup
  \renewcommand{\thefootnote}{}
  \footnotetext{Code available at: \url{https://github.com/liuktc/RefineCAM}}%
\endgroup

Some studies on attribution maps assess the quality of proposed methods through user studies \cite{achtibat2023attribution,e2024evaluating}, for example by asking human evaluators to judge how informative or accurate an attribution map is.
While effective, relying on users makes it difficult to fairly compare many attribution methods across different models and datasets.
Furthermore, users may be subject to expectation bias. More fundamentally, this reflects a problem in how the task is specified: the attribution map should reflect the model’s predictions rather than what the user expects.
For example, if a model has learned a shortcut or a spurious correlation \cite{geirhos2020shortcut} and the attribution map correctly highlights it, a user may judge it to be a “bad” attribution map, since it is not the map the user expects to see. As an example, given the task of classifying if an x-ray image contains a tumor or not, if all the true positive images contain a small cross at the bottom, the model will learn to classify given that cross. In this case, the user will evaluate the attribution map focused on the cross to be a bad attribution map, even though it correctly highlights the important pixels for the model prediction.

For this reason, quantitative evaluation of an attribution map is extremely important. 
However, the main problem with this approach is the absence of ground truth labels. This can be tackled in two ways. The first way is to leverage a synthetic dataset built in such a way as to have reliable ground-truth attribution maps \cite{hesse2023funnybirds,bohle2021convolutional}.
The second way is to use metrics that do not rely on a ground truth. 
When working with real-world datasets, the second option is the only available choice. For this reason, many metrics that do not need a ground truth have been developed. The metrics mainly rely on perturbation-based strategies, such as masking parts of the input image \cite{GradCAM,GradCAM++,AblationCAM,ScoreCAM}. The underlying assumption is that masking important regions should alter the model’s prediction, while masking irrelevant areas should have minimal effect. Unfortunately, while these metrics are intuitive, they can be easily fooled by trivial attribution maps (e.g., randomly uniform distributed), exposing the limits of such metrics (as we will later show in \Cref{sec:empirical_validation}).

We address these challenges by introducing a synthetic framework where class-background correlations are absent by design. In this way, we ensure a controlled environment that allows us to identify existing metrics that best reflect ground-truth alignment. Building on these insights, we propose ARCC, an aggregative metric that shows the highest correlation with attribution quality within our synthetic framework.

Beyond the challenge of evaluation, most existing attribution methods tend to produce attribution maps in the form of high-level blobs, with limited spatial precision. However, the level of detail of visual explanation is task dependent and for a range of critical tasks a fine-grained spatial precision is necessary. For instance, in medical imaging,  a dermatologist may need to assess the border irregularities of a skin lesion for signs of malignancy. In this case, a coarse heatmap that only highlights the general area of a mole, is fundamentally inadequate \cite{TowardMedicalXAI}. 
The standard approach in the literature of using the last convolutional layer before the classification head to generate the attribution maps \cite{GradCAM,GradCAM++,ScoreCAM} is motivated by the fact that the high-level, semantic features are captured by the deeper layers of the network. While shallower layers offer the potential for high-resolution explanations, they are often ignored because standard CAM methods typically produce noisy and less discriminative attribution maps from them (see \Cref{fig:RefineCAM_Metrics} for a qualitative example).
We therefore propose a simple methodology, denoted RefineCAM, which improves CAMs from shallow layers by leveraging deeper-layer attributions to produce cleaner, high-resolution explanations.

\noindent
Our main contributions are:

(1) We introduce ARCC, a novel composite metric that robustly detects incorrect attributions, addressing shortcomings of existing metrics identified in our review.

(2) We design a synthetic dataset with ground-truth attribution maps, providing a rigorous framework for quantitatively evaluating attribution metrics and revealing ARCC’s superior effectiveness and robustness.

(3) Finally, we propose RefineCAM, a general meta-method that leverages semantics from deeper layers to generate high-resolution, high-fidelity attribution maps from shallow layers for any CAM-based approach.

\section{Related Work}

Related work on visual explanations can be grouped into three threads: (i) evaluation metrics and benchmarking strategies for attribution quality, (ii) CAM and saliency methods that differ in how they compute class importance from activations or gradients, and (iii) methods that aim to increase spatial precision or refine attribution maps. We briefly summarize representative works in each thread and highlight the limitations that motivate our contributions.

(i) Evaluating explanations is challenging because ground truth is generally unavailable in real images. Perturbation-based metrics \cite{ROAD,GradCAM++} and composite scores \cite{ADCC} aim to measure faithfulness and compactness, but can be fooled by trivial maps (see \Cref{sec:empirical_validation} for a detailed discussion). Synthetic benchmarks such as FunnyBirds \cite{hesse2023funnybirds} or Grid Pointing game \cite{bohle2021convolutional} enable the evaluation of attribution maps by providing explicit ground-truth structures, allowing metrics to be validated under controlled conditions. Nevertheless, these prior benchmarks focus primarily on semantic parts rather than fine-grained, low-level detail. This motivates both our synthetic dataset design and our ARCC metric which combines rank-sensitive and magnitude-sensitive components.

(ii) CAM-style approaches produce attribution maps by combining activation maps with scalar or spatial weights. Gradient-based methods \cite{GradCAM++,GradCAM,simonyan2013deep,sundararajan2017axiomatic,LayerCAM} and perturbation-based approaches \cite{ScoreCAM,AblationCAM} differ mainly in whether they rely on gradients or model responses. A recurring trade-off is semantic fidelity versus spatial precision: deeper layers provide class semantics but at low spatial resolution, while shallow layers have higher resolution but are noisier. This gap highlights the need for high-resolution and semantically accurate attribution maps.

(iii) Several methods improve localization using multi-resolution inputs, as in MasterCAM \cite{masterCAM}, pixel-level gradients, as in Guided GradCAM \cite{GradCAM}, or spatially resolved gradient weighting over feature maps, as in LayerCAM \cite{LayerCAM}.
While these approaches tend to produce better-localized attribution maps at shallow layers compared to standard CAM methods, they do not exploit the hierarchical feature representations produced by the network across layers.
In contrast, RefineCAM introduces a simple multiplicative refinement that uses deep-layer attributions to denoise and sharpen high-resolution shallow-layer maps.

\section{Evaluating the Evaluators}
\label{sec:experiments}

\subsection{Review of metrics}
\label{sec:evaluation_metrics}

\paragraph{Notation}

Consider a neural network $Y=f(X)$ with input image $X\in\mathbb{R}^{3\times H\times W}$ and output class probabilities $Y$, where $f(X)^c$ denotes the probability of class $c$.
We denote by $M^c(X)\in[0,1]^{H\times W}$ a generic attribution map for image $X$ and class $c$, with larger values indicating greater pixel importance.
All evaluation metrics apply to such generic attribution maps and are independent of the attribution method.
For CAM-like methods, let $A_l\in\mathbb{R}^{C\times H'\times W'}$ be the activations of layer $l$, with channels $A_l^k\in\mathbb{R}^{H'\times W'}$.
CAM methods compute a class-specific attribution map as $L_l^c=\mathrm{ReLU}\!\left(\sum_k w_k^c A_l^k\right)$, where $w_k^c$ are class-specific weights \cite{CAM,GradCAM,ScoreCAM,AblationCAM}.
We denote the CAM output for image $X$ at layer $l$ and class $c$ as $CAM_l^c(X)=L_l^c$, which is a specific instance of $M^c(X)$.
All attribution maps are resized if needed and normalized to $[0,1]$ before computing any metric, yielding $M^c(X)\in[0,1]^{H\times W}$.
We denote the Hadamard product by $\odot$.

\paragraph{Average Drop}
The Average Drop \cite{GradCAM++} metric evaluates the faithfulness of an attribution map by measuring the reduction in model confidence for a specific class when the input is masked by the explanation map via element-wise multiplication, i.e., $X \odot L_l^c$, so that only the regions emphasized by the attribution are preserved while the rest of the image is suppressed.
A smaller drop indicates that the explanation map effectively captures the regions most pertinent to the model's decision.
It is formulated as
\begin{equation}
    \text{AD}(X,L_l^c) = \frac{\max(0, f(X)^c - f(X \odot L_l^c)^c)}{f(X \odot L_l^c)^c}\mbox{.}
\end{equation}
The $\max(0, \cdot)$ function accounts for instances where showing only the explanation map might unexpectedly increase confidence.
The Average Drop metric exhibits a notable limitation regarding its sensitivity to noise. Erroneously including additional, irrelevant regions in an attribution map (i.e., increasing the heatmap's active area) is unlikely to decrease the class probability $f(X)^c$, as more of the original image content is revealed. Consequently, the metric fails to penalize explanations that are not minimal or that contain noise. This characteristic is demonstrated by the fact that a trivial attribution map composed only of ones yields the best Average Drop score.

\paragraph{Complexity}

Complexity \cite{ADCC} measures the fraction of the attribution that is active, defined as:
\begin{equation}
    \text{Cmx}(L_l^c) = \| \mathit{flatten}(L_l^c) \|_1 
\end{equation}
where $\| \cdot \|_1$ denotes the L1-norm of a vector and  $\mathit{flatten}(\cdot)$ denotes the 1D flattened version of the matrix $L_l^c$. The intuition is that a good attribution map should not highlight too much of the input image, but should be as small as possible. Note that a trivial attribution map composed only of 0s obtains the best score for Complexity.

\paragraph{Coherency} 

The Coherency metric \cite{ADCC} evaluates the stability and self-consistency of an attribution map. If an attribution map correctly highlights the important features, then re-computing the map on an image that has been masked by this very explanation should ideally produce a similar attribution map.

This stability is quantified by computing the Pearson correlation coefficient between the original attribution map and the new map generated from the self-masked image. A high degree of correlation suggests that the explanation is self-consistent and not heavily reliant on contextual information outside of the primary salient regions. The metric is formally defined as:
\begin{equation}
    \text{Chn}(X, L_l^c) = \frac{\text{Cov} (\text{CAM}_l^c(X \odot L_l^c)), L_l^c)}{\sigma_{\text{CAM}_l^c(X \odot L_l^c)}  \sigma_{L_l^c}}
\end{equation}
where $\text{Cov}(\cdot, \cdot)$ denotes the covariance, and $\sigma_L$ represents the standard deviation of the flattened attribution map $L$. A Coherency score close to 1 indicates a highly stable and self-consistent explanation.

\paragraph{ADCC}
Single-component evaluation metrics can be vulnerable to trivial attribution maps; for example, an all 1s map may maximize Average Drop but will have minimal score on Complexity. To provide a more comprehensive evaluation that penalizes such naive explanations, the ADCC metric was introduced \cite{ADCC}. ADCC aggregates three complementary metrics -- Average Drop, Coherency, and Complexity -- using the harmonic mean. It is defined as:
\begin{align}
    \text{ADCC}(X, L_l^c) = 3  \left(
    \frac{1}{\text{Chn}(X, L_l^c)}\right. 
    \left.+\frac{1}{1-\text{Cmx}(L_l^c)}+
    \frac{1}{1-\text{AD}(X, L_l^c)}
    \right) ^{-1}
\end{align}
Despite its intent, we note here that ADCC remains susceptible to certain types of trivial attributions. As we demonstrate in our experiments (see \Cref{sec:empirical_validation}), it is possible to construct naive attribution maps that achieve favorable scores for both Complexity and Average Drop simultaneously. This manipulation results in a misleading high ADCC score for a trivial explanation, showing a vulnerability in this composite metric and motivating our proposal for the more robust ARCC metric.

\paragraph{ROAD (Remove And Debias)}
The ROAD (Remove And Debias) framework \cite{ROAD} involves progressively perturbing the input image based on the attribution map in two complementary orders:
Most Relevant First (MoRF), which perturbs pixels in descending order of importance, and Least Relevant First (LeRF), which uses ascending order. The intuition is that, for a faithful explanation, MoRF perturbations should cause a sharp drop in model confidence, while LeRF perturbations should have a minimal impact. To avoid artifacts from simple pixel replacement, ROAD uses a Noisy Linear Imputation \cite{ROAD} method, which blends perturbed regions with their surroundings.
A final score is computed by averaging the confidence differences between LeRF and MoRF perturbations over a set of pixel fractions $K$.
A higher score indicates a more faithful explanation. The score is defined as:
\begin{equation}
     \text{ROAD}(X, L_l^c) = \frac{1}{|K|} \sum_{k \in K} \left( f(X'_{\text{LeRF}, k})^c - f(X'_{\text{MoRF}, k})^c \right)
\end{equation}
where $X'_{\text{LeRF}, k}$ denotes the input image $X$ with top $k$ less important pixels perturbed, $X'_{\text{MoRF}, k}$ denotes $X$ with the top $k$ most important pixels perturbed.
In our experiments, we set $K = \{0.2, 0.4, 0.6, 0.8\}$. 
Although ROAD represents a significant improvement over Average Drop (as shown in \Cref{ssec:empirical_results}), it evaluates only the order of attribution scores. This means that ROAD is sensitive to the ranking of the most and least relevant pixels, but not to the specific values themselves. This means that if the values in the attribution map change but their relative ordering remains the same, the ROAD score remains unchanged (see \Cref{fig:road_problem}).

\subsection{Proposed ARCC metric}
Although ADCC mitigates some metric limitations, its reliance on Average Drop leaves it vulnerable to some trivial baselines (more details on \Cref{sec:empirical_validation}). To address this, we propose ARCC, a new metric that builds on the superior robustness of ROAD. ARCC is computed by substituting Average Drop with ROAD in the harmonic mean formulation of ADCC.
While ROAD is more computationally expensive than Average Drop, we prioritize it due to its superior robustness and faithfulness.
Furthermore, ROAD is combined with Coherency and Complexity metrics. Since these are sensitive to the absolute values of the attribution map, their combination ensures a more robust and comprehensive evaluation.
The ARCC metric is defined as:
\begin{align}
    \text{ARCC}(X, L_l^c) = 3  \left(
    \frac{1}{\text{Chn}(X, L_l^c)} \right.
    \left. + \frac{1}{1-\text{Cmx}(L_l^c)} +
    \frac{1}{\text{ROAD}(X, L_l^c)}
    \right) ^{-1}
\end{align}

\subsection{Empirical validation}
\label{sec:empirical_validation}

Based on the intuitions discussed in \ref{sec:evaluation_metrics}, where we highlighted the limitations of certain evaluation metrics, we conducted an empirical evaluation by using trivial CAMs. Specifically, we compared GradCAM++ \cite{GradCAM++}, LayerCAM \cite{LayerCAM} and ScoreCAM \cite{ScoreCAM} with 3 intentionally trivial attribution maps: \textit{RandomCAM}, in which each pixel is sampled uniformly from the interval $[0, 1]$; \textit{HalfCAM}, which assigns a value of  $0.5$ to all but one pixel set to $0$ and another set to $1$; and \textit{All1sCAM}, which assigns all values to $1$ except for a single pixel to $0$.

\begin{figure}
    \centering
    \includegraphics[width=.8\linewidth]{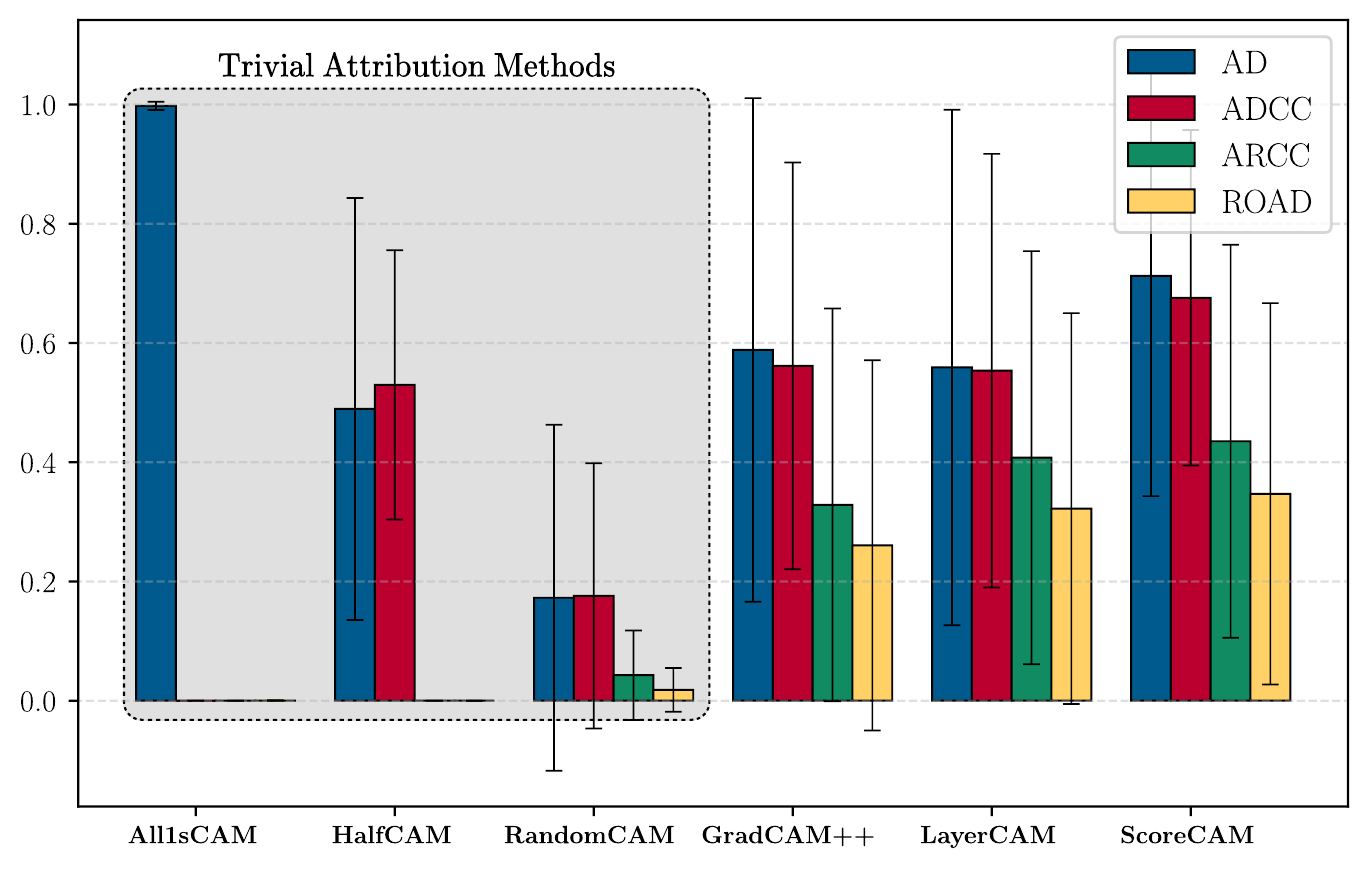}
    \caption{Results of various metrics computed on a random sample of $1000$ images from ImageNet. The results are computed on ResNet18, VGG11, Swin-Tiny and ConvNeXt-Tiny. For each attribution method we report the mean value of each metric and its standard deviation. In contrast to AD and ADCC, only ARCC and ROAD demonstrate correct discriminative behavior, penalizing all trivial methods with low scores while rewarding non-trivial methods with high scores.}
    \label{fig:trivial_results}
\end{figure}

\Cref{fig:trivial_results} presents the results, where the different attribution methods have been computed on a random subsample of 1000 ImageNet \cite{russakovsky2015imagenet} images. All experiments were conducted on ResNet18 \cite{ResNet}, VGG11 \cite{vgg}, Swin-Tiny \cite{Swin}and ConvNeXt-Tiny \cite{ConvNeXt}. We observe that HalfCAM fools both Average Drop and ADCC, producing values comparable with non-trivial CAMs. In contrast, ARCC and ROAD correctly assign low scores to these trivial CAMs, validating their theoretical robustness and reliability towards trivial attribution maps.

\subsection{Synthetic Dataset for Controlled Evaluation}
\label{ssec:synthetic_dataset}

Evaluating attribution methods quantitatively requires access to a ground-truth explanation. In real-world datasets, such ground truth is fundamentally ill-defined. Explanations are inherently model-dependent, reflecting the internal logic or biases of the trained network, which are not known a priori. In practice, a model may learn to rely on features or shortcuts that differ significantly from human intuition, making it impossible to define a ground truth explanation in the general case. 

\begin{figure}[!t]
    \centering
    \includegraphics[width=.8\linewidth]{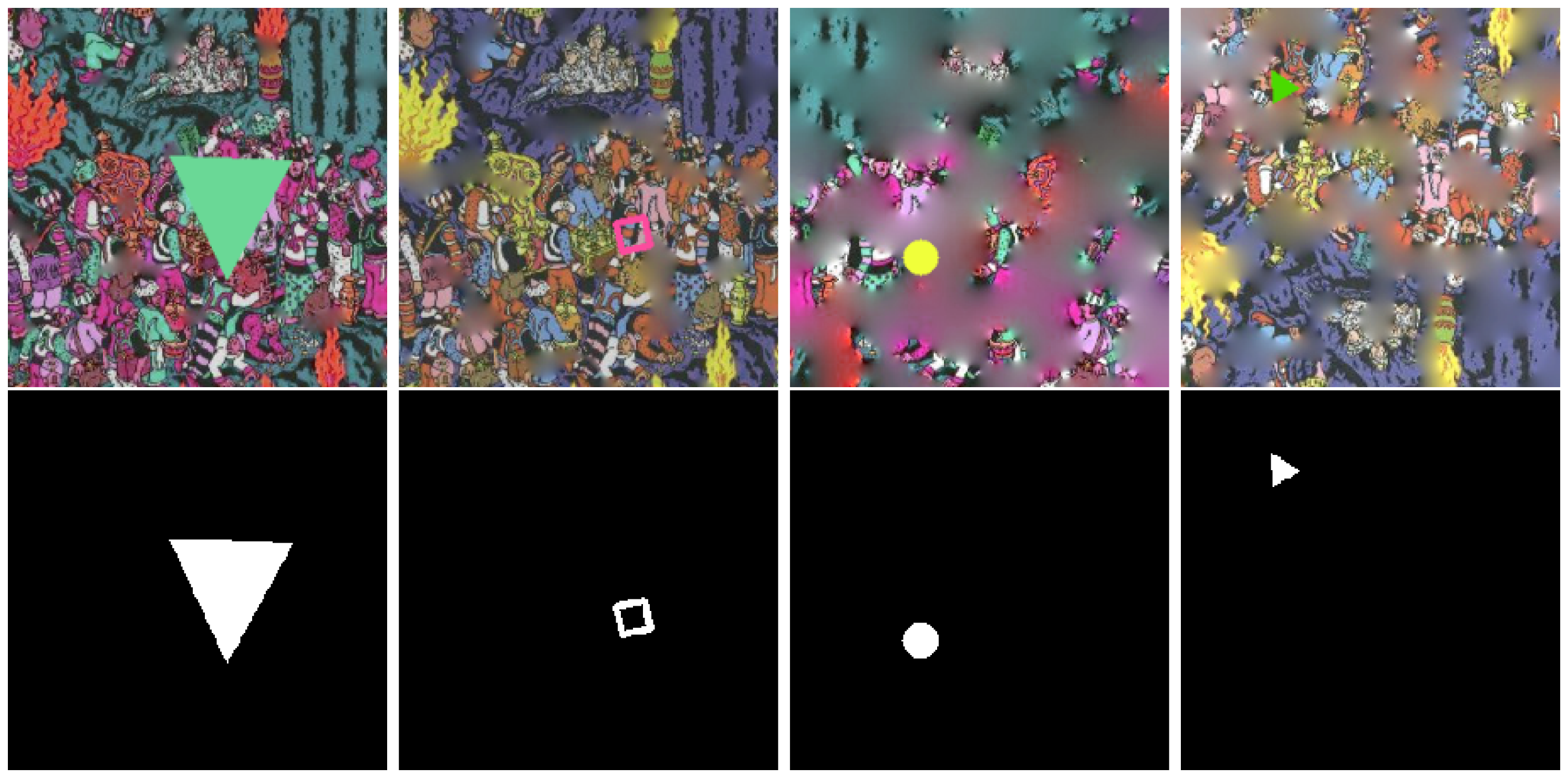}
    \caption{Example images (top row) from our proposed synthetic dataset paired with the corresponding ground truth attribution map (bottom row). }
    \label{fig:synth_examples}
\end{figure}

To obtain ground-truth attribution maps, we define a synthetic dataset of 6 different geometric shapes (circles, squares, and triangles, both filled and empty) overlaid on complex backgrounds from the \textit{Where’s Waldo?} series \cite{whereswaldy_dataset}. Each shape is randomly colored, rotated, scaled and positioned to have as much variability as possible. In this way, we create a dataset containing also very small objects, specifically to challenge low resolution, blob-like attribution maps.
Some sampled examples from our synthetic dataset are shown in \cref{fig:synth_examples}.
To guarantee that classification by the tested models only depends on the foreground pixels of the geometric figure, we use the same exact background distribution for all the 6 different classes. The background images used in the train set are different from the ones used in the test set. This creates an important difference with real datasets like ImageNet, where many classes frequently appear only with specific backgrounds. For example, certain bird species may be displayed exclusively against particular backgrounds like trees.
In this case it is impossible to establish a ground-truth attribution map by segmenting only the object of the class, like the bird, because the model may base its classification also on background cues. In our case, background features are clearly not predictive of the class, hence including them in the attribution is definitely an error. Moreover, all the models used in our experiments achieve around 99\% test accuracy, indicating that classification is driven by foreground shape and not background artifacts.
Under these conditions, we can reasonably assume that the model's decision regions align with the ground truth attribution map.

\subsection{Synthetic Dataset - Experimental Results}
\label{ssec:empirical_results}
To quantitatively validate and rank metrics for CAM evaluation, we run 5 CAM methods (GradCAM++ \cite{GradCAM++}, LayerCAM \cite{LayerCAM}, ScoreCAM \cite{ScoreCAM}, ShapleyCAM \cite{ShapleyCAM} and GuidedGradCAM \cite{GradCAM}) and 5 other attribution methods (LIME \cite{ribeiro2016should}, DeepLift \cite{DeepLift}, InputXGradient \cite{DeepLift}, IntegratedGrads \cite{sundararajan2017axiomatic} and KernelShap \cite{SHAP}) on our synthetic dataset described in \Cref{ssec:synthetic_dataset} using ResNet18 \cite{ResNet}, VGG11 \cite{vgg}, Swin-Tiny \cite{Swin} and ConvNeXt-Tiny \cite{ConvNeXt}. 
All the CAM methods are applied at the end of the shallowest stage of each  model.
We then  assess the quality of the attribution maps by computing the Cosine similarity as $sim(L, L_{\text{GT}}) = \frac{L \cdot L_{\text{GT}}}{\| L \| \|L_{\text{GT}}\|}$, where $L$ and $L_{\text{GT}}$ are the predicted and ground-truth attribution maps, respectively.
To further validate the results, we also use the same models and attribution methods on FunnyBirds \cite{hesse2023funnybirds}, a synthetic dataset built for evaluating attribution methods by calculating importance scores based on removing object parts. While a full description of the FunnyBirds score is beyond the scope of this paper, we outline its main components below. The FunnyBirds score is the average of the following: \textit{Completeness} measures how well an explanation captures necessary objects parts; \textit{Correctness} computes the correlation between assigned part importance and actual model output change; \textit{Contrastivity} evaluates if explanations correctly highlight class-specific parts when the target class is changed. Note that this score can only be computed on the FunnyBirds dataset. 

\begin{figure*}[!h]
    \centering
    \begin{subfigure}{0.49\linewidth}
        \centering
        \includegraphics[width=\linewidth]{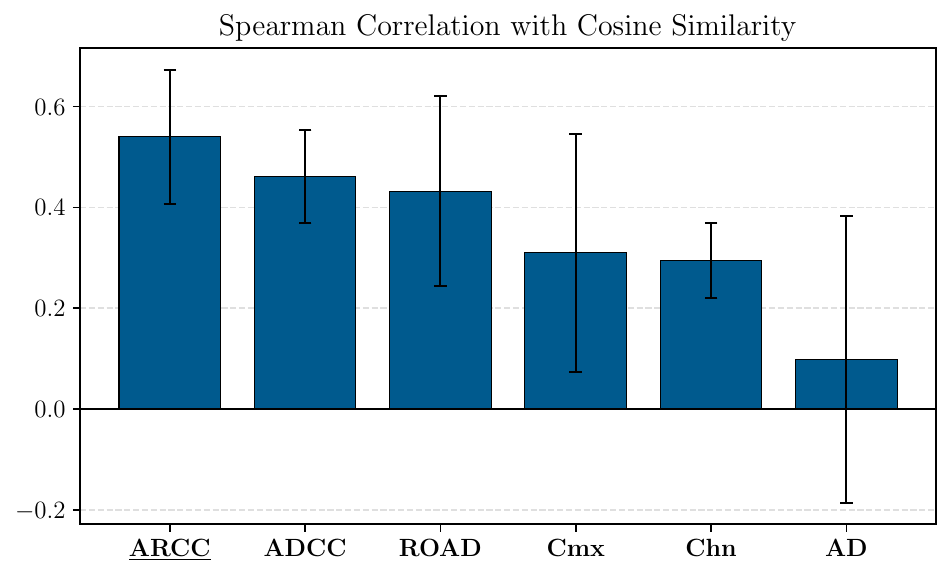}
        \caption{Correlation on our synthetic dataset}
        \label{fig:corr_att_a}
    \end{subfigure}
    \hfill
    \begin{subfigure}{0.49\linewidth}
        \centering
        \includegraphics[width=\linewidth]{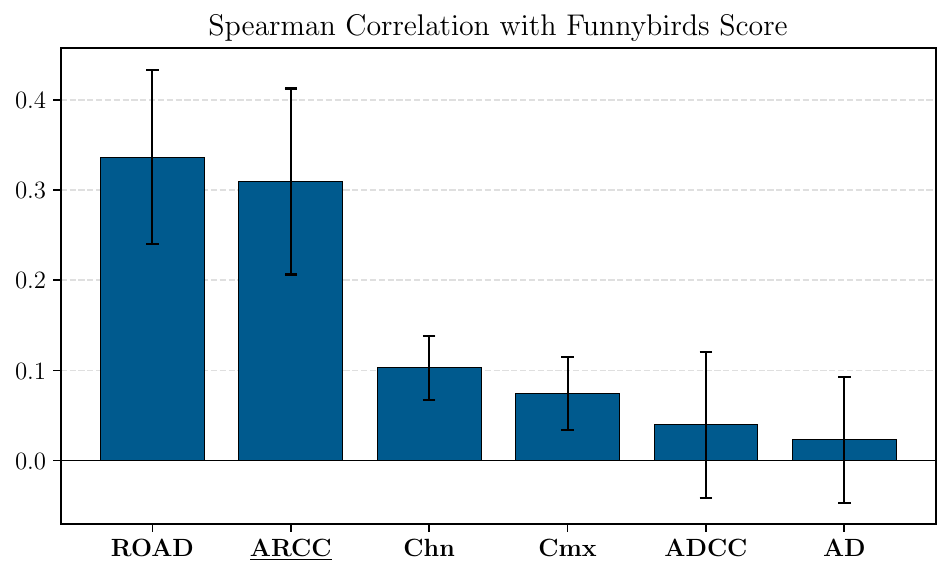}
        \caption{Correlation on the Funnybirds dataset}
        \label{fig:corr_att_b}
    \end{subfigure}
    \caption{Pearson correlation between the studied metrics and: (a) Cosine Similarity (our synthetic dataset) and (b) FunnyBirds Score (FunnyBirds \cite{hesse2023funnybirds} dataset). Results for each metric are aggregated across the $4$ tested models. For each metric we report the mean and standard deviation across models of the Spearman Correlation.}
    \label{fig:corr_att}

    \centering
    \includegraphics[width=0.75\linewidth]{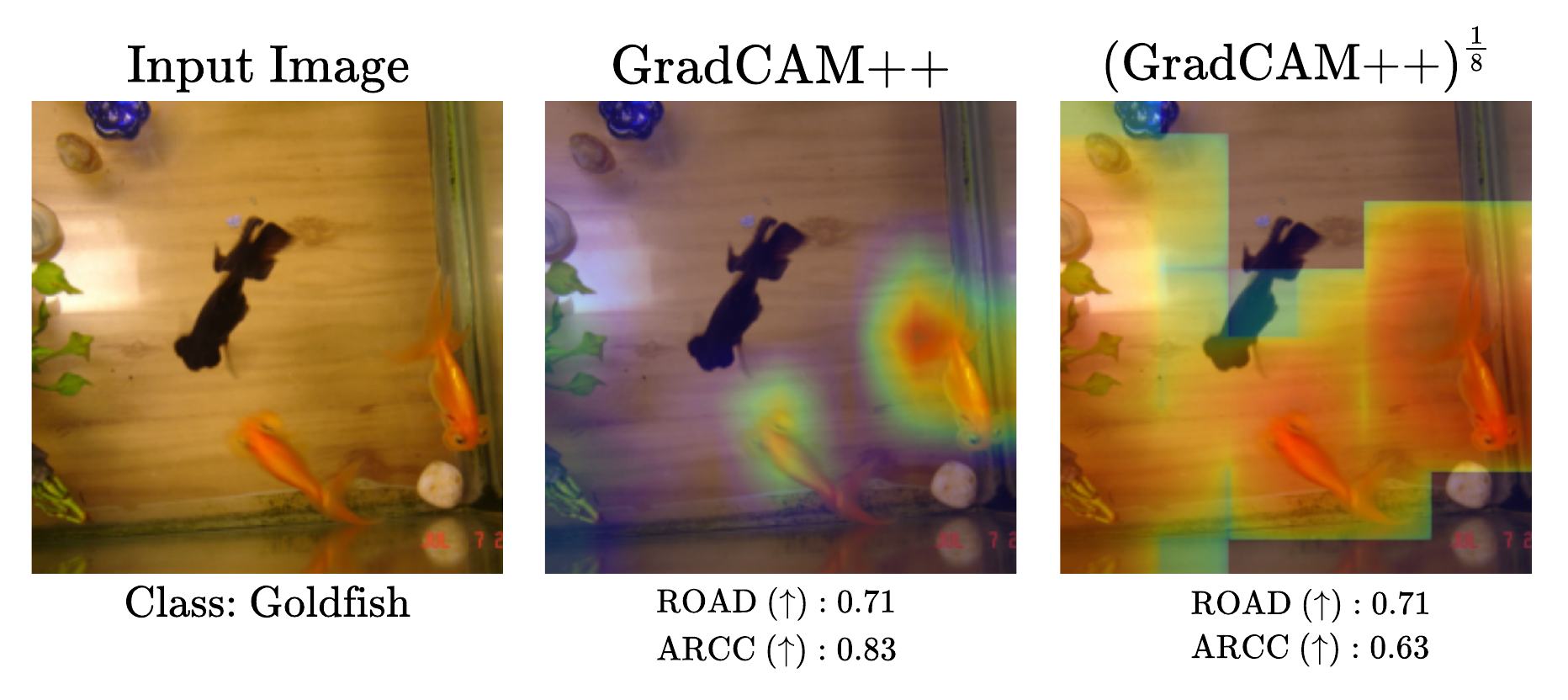}
    \caption{Comparison of GradCAM++ (center) and its $\frac{1}{8}$ power transformation (right). The ROAD score remains unchanged ($0.71$) because monotonic transformations preserve the pixel rank ordering, which is the sole factor ROAD measures. In contrast, ARCC reacts to the change in magnitude ($0.83 \rightarrow 0.63$), showing its superior sensitivity to attribution map changes.}
    \label{fig:road_problem}
\end{figure*}

To compare the metrics, we compute the Pearson correlation coefficient between each metric's score and the two ground-truth scores: Cosine Similarity on the synthetic dataset and the FunnyBirds Score on the FunnyBirds dataset. The idea is that these two scores effectively capture the \textit{true} quality of the attribution map. For this reason, a higher correlation with these scores indicates a more reliable and effective evaluation metric. The results across both datasets are presented in \Cref{fig:corr_att}.
On the synthetic dataset, ARCC outperforms all other metrics. On FunnyBirds, ARCC and ROAD achieve very similar results and, when accounting for the standard deviation, their performance is statistically comparable. The main advantage of ARCC over ROAD lies in its sensitivity to the magnitude of the attribution values. While ROAD considers only the ordering of pixels (as illustrated in \Cref{fig:road_problem}), ARCC takes the attribution values themselves into account, enabling it to distinguish between overly broad activations and more localized ones.
Interestingly, Average Drop (AD), although widely adopted, consistently yields the lowest correlation across all evaluated metrics, suggesting limited reliability.

\section{Refine-CAM}

\subsection{Layer Selection in CAM Methodologies}
\label{ssec:layer_selection}

\begin{figure*}[t]
    \centering
    \includegraphics[width=1\linewidth]{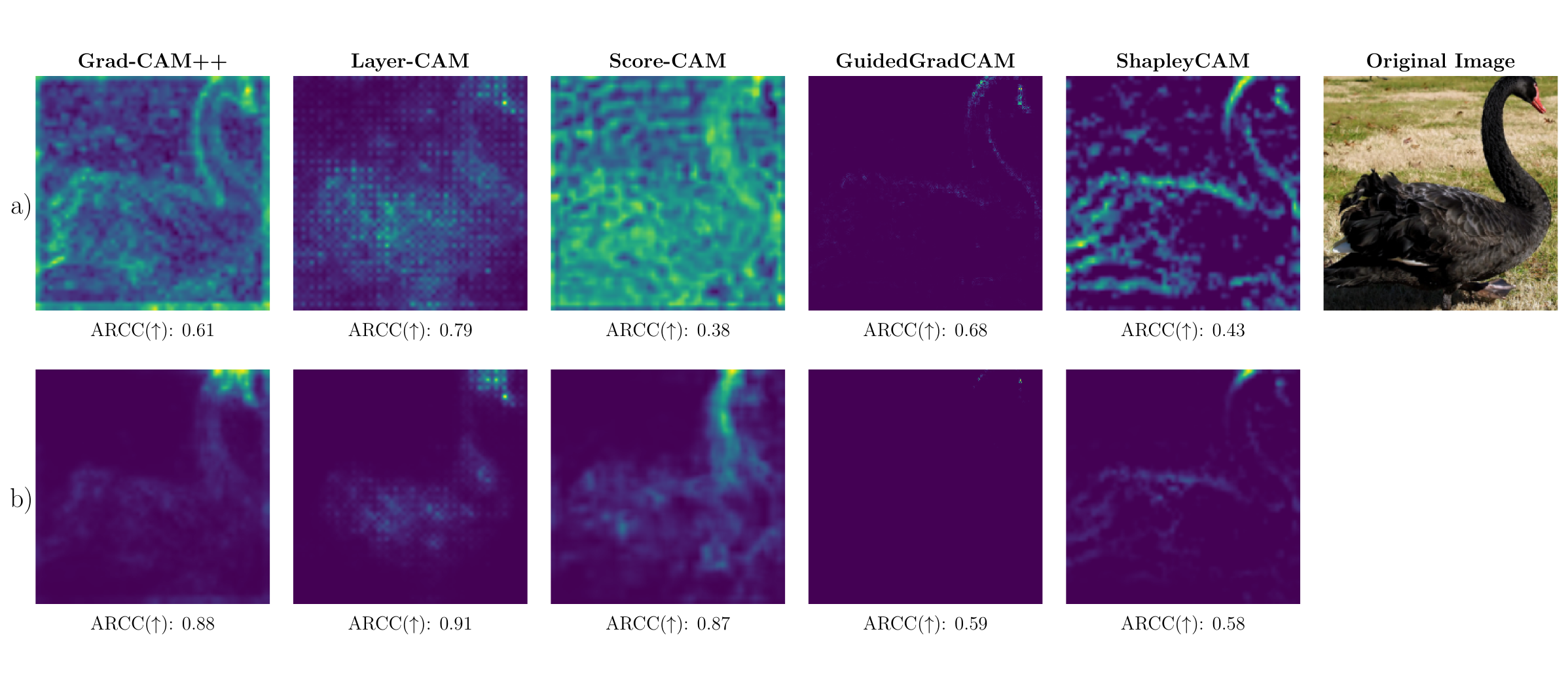}
    \caption{Visual comparison of attribution maps generated on ImageNet. (a) shows the noisy maps from the original methods applied to a shallow layer. (b) shows the refined maps using RefineCAM applied to the same shallow layer. The attribution maps are computed using ResNet18.} 
    \label{fig:RefineCAM_Metrics}
\end{figure*}

A common design choice in CAM methods is to extract attribution maps from the last convolutional layer before the classification head\cite{GradCAM,GradCAM++,ScoreCAM}. This is motivated by the fact that deeper layers encode high-level semantic features that are closely aligned with class-specific decision boundaries\cite{zeiler2014visualizing}. However, these deep feature maps typically have low spatial resolution (e.g., $7\times 7$ or $14\times 14$ for standard $224 \times 224$ inputs), which needs aggressive upsampling to match input dimensions. This interpolation process introduces artifacts and obscure fine-grained details.
In contrast, shallower layers produce higher-resolution activation maps (e.g., $56\times 56$) that capture low-level visual primitives such as edges and textures. These finer details are critical for applications requiring precise spatial attribution. Nevertheless, attribution maps derived directly from shallow layers are often noisy and less class-discriminative, leading to limited practical use in CAM-based techniques (see \Cref{fig:RefineCAM_Metrics}(a) for a qualitative example).

This trade-off between semantic quality in deep layers and spatial precision in shallow layers motivates our proposed method. RefineCAM leverages deep-layer semantic information to guide and refine the high-resolution but noisy attributions from shallow layers, combining the strengths of both. 

\subsection{Methodology}

Given a list of $N$ layers $\mathcal{L}=\{l_1, l_2, \dots,l_N\}$, ordered from shallow to deep, within a deep neural network, and a selected layer $l \in \mathcal{L}$, our proposed meta-method, RefineCAM, for an arbitrary CAM-like attribution technique, denoted as  Your-CAM, is defined as:
\begin{equation}\label{eqn:MyMethod}
    L_{l, \text{Refine-CAM}}^c = \prod_{l' \in \mathcal{L}, l' \geq l} L_{l', \text{Your-CAM}}^c
\end{equation}
where $l' \geq l$ means that $l'$ represents a layer situated deeper or at the same depth in the network architecture than layer $l$. The product operation between attribution maps is defined as element-wise multiplication.
Prior to this operation, all attribution maps $L_l^c$ are resized to match the spatial dimensions of the input image via bilinear upsampling. After the multiplication, the resulting map is normalized in the range $[0,1]$.
Due to its multiplicative nature, RefineCAM operates purely as a refinement step, therefore if the original attribution maps omit import regions, it is unable to recover them. 

By exclusively mixing attribution maps derived from deeper layers ($l' \geq l$), our approach leverage the high-level semantic feature localization capabilities inherent to deeper layers to mitigate noise present in attribution maps generated from shallower layers. This multiplicative combination can be interpreted as a fuzzy logical \texttt{AND} operation, wherein pixels retaining high importance values after the multiplication have consistent significance across all deeper layers considered.
Although we explored more complex aggregation operators, we found element-wise multiplication to be the most effective at refining attribution maps. We provide an  ablation study on this choice in \Cref{ssec:ablation_aggregation}.

\subsection{RefineCAM - Experimental Results}

We quantitatively assess the performance of RefineCAM using ARCC. Specifically, we focus on comparisons between 5 standard CAM-based techniques (Grad-CAM++\cite{GradCAM++}, LayerCAM\cite{LayerCAM}, ScoreCAM\cite{ScoreCAM}, ShapleyCAM\cite{ShapleyCAM} and GuidedGradCAM \cite{GradCAM}) and their RefineCAM counterparts when applied to the shallowest convolutional layer of the network ($l = l_1$ in \cref{eqn:MyMethod}). For a more complete comparison we also include the following attribution methods: LIME\cite{ribeiro2016should}, DeepLift\cite{DeepLift}, InputXGradient \cite{DeepLift}, IntegratedGrads\cite{sundararajan2017axiomatic} and KernelShap\cite{SHAP}. Since these methods are layer-agnostic and do not depend on feature maps from a specific network layer, RefineCAM is not applicable. These experiments are carried out on a random subset of $1000$ images from ImageNet.
As shown in \Cref{tab:refine_cam}, applying RefineCAM generally improves upon the base CAM methods across architectures. In particular, RefineCAM achieves the strongest ARCC performance among CAM-based approaches on three of the four evaluated models, demonstrating its effectiveness across diverse network designs.
\Cref{fig:RefineCAM_Metrics} offers a qualitative visualization of attribution maps produced by both original and RefineCAM-enhanced methods.

\begin{table}[h]
\centering
\small
\begin{tblr}{
  colspec = {l *{8}{Q[c, wd=1.1cm]}},
  row{1} = {font=\bfseries},
  rows = {rowsep=0pt, abovesep=1pt, belowsep=1pt},
  leftsep = 1pt,
  rightsep = 1pt,
  column{3,5,7} = {rightsep=1pt},
  hline{1,3,Z} = {wd=0.8pt},
  hline{8} = {wd=0.2pt,dotted},
  cell{3-Z}{1} = {font=\footnotesize},
}
XAI Method &
  \SetCell[c=2]{c} VGG11 & & 
  \SetCell[c=2]{c} ResNet18 & & 
  \SetCell[c=2]{c} Swin-Tiny & & 
  \SetCell[c=2]{c} ConvNext-Tiny \\
& Base & \textit{OURS} & Base & \textit{OURS} & Base & \textit{OURS} & Base & \textit{OURS} \\
GradCAM++ &
  \snum{.22}{.31} & \snum{.27}{.36} &
  \snum{.37}{.29} & \snum{.51}{.31} &
  \snum{.15}{.22} & \snum{.07}{.18} &
  \snum{.47}{.32} & \SetCell{bg=gray!20}\snumbf{.65}{.30} \\
LayerCAM &
  \snum{.28}{.36} & \snum{.29}{.37} &
  \snum{.52}{.30} & \snum{.55}{.31} &
  \snum{.30}{.28} & \snum{.27}{.30} &
  \snum{.58}{.28} & \snum{.64}{.25} \\
ScoreCAM &
  \snum{.24}{.32} & \snum{.29}{.37} &
  \snum{.37}{.27} & \snum{.54}{.30} &
  \snum{.41}{.31} & \SetCell{bg=gray!20}\snumbf{.50}{.32} &
  \snum{.34}{.32} & \snum{.35}{.35} \\
ShapleyCAM &
  \snum{.44}{.35} & \snum{.49}{.37} &
  \snum{.42}{.30} & \SetCell{bg=gray!20}\snumbf{.57}{.27} &
  \snum{.28}{.28} & \snum{.09}{.20} &
  \snum{.42}{.34} & \snum{.48}{.35} \\
GuidedGradCAM &
  \snum{.42}{.39} & \snum{.37}{.40} &
  \snum{.51}{.33} & \snum{.44}{.37} &
  \snum{.28}{.28} & \snum{.19}{.28} &
  \snum{.39}{.31} & \snum{.36}{.32} \\
LIME &
  \snum{.19}{.28} & \NA &
  \snum{.14}{.24} & \NA &
  \snum{.26}{.30} & \NA &
  \snum{.22}{.27} & \NA \\
DeepLift &
  \SetCell{bg=gray!20}\snumbf{.54}{.24} & \NA &
  \snum{.37}{.27} & \NA &
  \snum{.07}{.14} & \NA &
  \snum{.14}{.20} & \NA \\
InputXGradient &
  \snum{.33}{.28} & \NA &
  \snum{.23}{.25} & \NA &
  \snum{.14}{.21} & \NA &
  \snum{.14}{.20} & \NA \\
IntegratedGrads &
  \snum{.44}{.27} & \NA &
  \snum{.33}{.27} & \NA &
  \snum{.17}{.23} & \NA &
  \snum{.18}{.22} & \NA \\
KernelShap &
  \snum{.25}{.21} & \NA &
  \snum{.21}{.20} & \NA &
  \snum{.18}{.19} & \NA &
  \snum{.18}{.19} & \NA \\
\end{tblr}
\vspace{0.2cm}
\caption{ARCC ($\uparrow$) of RefineCAM applied to 4 different models and 10 different attribution methods on a random subset of $1000$ images from ImageNet. The column \textit{OURS} denotes the application of RefineCAM to the Base attribution method. Results for CAM methods are computed on the shallowest layer of the network. The final five rows include standard attribution methods (e.g., LIME, DeepLift) to provide a more complete comparison against established baselines. N.A. is reported for these entries under \textit{OURS} column, as these methods do not utilize the layer selection hyperparameter required by RefineCAM. The best result for each model is highlighted in bold.}
\label{tab:refine_cam}
\end{table}

\subsection{Ablation Study on Aggregation Methods}
\label{ssec:ablation_aggregation}

Our proposed methodology utilizes element-wise multiplication to aggregate attribution maps from different layers. To validate this design choice, we conducted an ablation study comparing it against other aggregation operators. 
We evaluated two alternatives:

\begin{itemize}
    \item \textbf{Geometric mean:} This operator computes the n-th root of the element-wise product of $N$ attribution maps. It represents a less stringent aggregation than direct multiplication, potentially preserving features with lower saliency scores.

    \begin{equation}\label{eqn:MyMethod_geometric}
        L_{l, \text{Refine-CAM}}^c = \sqrt[\leftroot{-2}\uproot{2}N]{\prod_{l' \in \mathcal{L}, l' \geq l} L_{l', \text{Your-CAM}}^c}
    \end{equation}

    \item \textbf{Inverse Exp mean:} This operator computes the generalized $f$-mean\cite{kolmogorov1930notion} with $f(x)=e^{\frac{1}{x}}$ of the various attributions.

    \begin{equation}\label{eqn:MyMethod_inverse_exp}
        L_{l, \text{Refine-CAM}}^c = {\ln \left(  \displaystyle \frac{1}{N} \displaystyle \sum_{l' \in \mathcal{L}, l' \geq l} \exp\left(\frac{1}{L_{l', \text{Your-CAM}}^c}\right)\right)}^{-1}
    \end{equation}
\end{itemize}

\begin{table}[h]

\begin{tabularx}{\linewidth}{l *{3}{>{\centering\arraybackslash}X}}
\toprule
\textbf{XAI Method} & \textbf{Inverse Exp mean} & \textbf{Geometric mean} & \textbf{Multiplication} \\
\midrule
GradCAM++ & $\text{\normalsize .57}\text{\small $\pm$}\text{\small .30}$ & $\text{\normalsize .57}\text{\small $\pm$}\text{\small .29}$ & $\textbf{\text{\normalsize .63}\text{\small $\pm$}\text{\small .33}}$ \\
Guided GradCAM & $\text{\normalsize .07}\text{\small $\pm$}\text{\small .21}$ & $\textbf{\text{\normalsize .15}\text{\small $\pm$}\text{\small .29}}$ & $\text{\normalsize .12}\text{\small $\pm$}\text{\small .26}$ \\
LayerCAM & $\text{\normalsize .68}\text{\small $\pm$}\text{\small .27}$ & $\text{\normalsize .70}\text{\small $\pm$}\text{\small .25}$ & $\textbf{\text{\normalsize .74}\text{\small $\pm$}\text{\small .26}}$ \\
ScoreCAM & $\text{\normalsize .65}\text{\small $\pm$}\text{\small .26}$ & $\text{\normalsize .64}\text{\small $\pm$}\text{\small .25}$ & $\textbf{\text{\normalsize .70}\text{\small $\pm$}\text{\small .27}}$ \\
ShapleyCAM & $\text{\normalsize .25}\text{\small $\pm$}\text{\small .34}$ & $\text{\normalsize .31}\text{\small $\pm$}\text{\small .35}$ & $\textbf{\text{\normalsize .32}\text{\small $\pm$}\text{\small .36}}$ \\
\bottomrule
\end{tabularx}
\vspace{0.2cm}
\caption{Comparison of different mixing methods for RefineCAM on various attribution methods. The table reports the mean and std of ARCC $(\uparrow)$ obtained by evaluating 100 randomly sampled images from ImageNet with the 4 different tested models. The best result for each attribution method is highlighted in bold.}
\label{tab:xai_aggregation_comparison}

\end{table}

In \Cref{tab:xai_aggregation_comparison} we present the comparative performance of these three aggregation strategies, evaluated using our primary metrics, ARCC. The results demonstrate that standard element-wise multiplication consistently outperforms the other approaches. We hypothesize that multiplication acts as a more effective noise filter. By mimicking more closely a logical \texttt{AND} operator on saliency, it requires a region to be deemed important across multiple layers to be retained in the final map.

\section{Conclusions}

The development of robust evaluation methods for attribution maps is challenged by the absence of ground truth and the limitations of existing metrics. To address this, we introduced a synthetic dataset with pixel-level ground-truth attributions, enabling rigorous comparisons among metrics. Our contribution, ARCC, is a novel composite metric that demonstrated to consistently outperform prior approaches in identifying faithful explanations. Furthermore, we proposed RefineCAM, a meta-method that leverages deeper semantic information to refine shallow-layer CAM attributions, yielding high-resolution and faithful explanations. These advancements collectively provide a more principled framework for both evaluating and generating visual explanations.

While our synthetic dataset provided the ground truth necessary for rigorous metric evaluation, it currently lacks the semantic and textural variety of natural scenes. Addressing this limitation is a priority for future work. We propose expanding the benchmark to include diverse object types, complex backgrounds, and occlusions, ensuring the data better reflects real-world distributions without sacrificing ground-truth reliability.



\bibliographystyle{splncs04}
\bibliography{egbib}

\end{document}